\documentclass[letterpaper, 10 pt, conference]{ieeeconf}  

\IEEEoverridecommandlockouts                              

\usepackage{graphics} 
\usepackage{graphicx}
\usepackage{mathptmx} 
\usepackage{times} 
\usepackage{amsmath,bm} 
\usepackage{amsfonts,amssymb} 
\usepackage{colortbl}
\usepackage[table]{xcolor}
\usepackage{booktabs}
\usepackage{multirow}
\usepackage{arydshln}
\usepackage{tabularx}
\usepackage{threeparttable} 
\usepackage{mathrsfs}
\usepackage{pifont}
\usepackage{urwchancal}
\usepackage{url}
\usepackage{placeins}
\usepackage{hyperref}
\usepackage{makecell}

\usepackage{cuted}
\usepackage{capt-of}

\newcommand{\Fig}[1]{Fig. \ref{#1}}
\newcommand{\eeq}{\end{equation}}
\newcommand{\beq}{\begin{equation}}

\newcommand{\Tab}[1]{Tab. \ref{#1}}

\newcommand{\MethodName}[1]{Hier-SLAM}
\newcommand{\bestcolor}[1]{\cellcolor{green!30}\textbf{#1}}
\newcommand{\bestcolornob}[1]{\cellcolor{green!30}{#1}}
\newcommand{\secondcolor}[1]{\cellcolor{yellow!30}{#1}}

\title{\LARGE \bf
Hier-SLAM: Scaling-up Semantics in SLAM with a Hierarchically Categorical Gaussian Splatting

}

\author{ Boying Li$^{1*}$, Zhixi Cai$^{1}$, Yuan-Fang Li$^{1}$, Ian Reid$^{2}$, and Hamid Rezatofighi$^{1}$
\thanks{$^{1}$ Faculty of Information Technology, Monash University, Australia. $^{2}$ Mohamed bin Zayed University of Artificial Intelligence, United Arab Emirates. $^*$ Corresponding author: Boying Li ({\tt\small boying.li@monash.edu)}
}
\thanks{This work is supported by the DARPA Assured Neuro Symbolic Learning and Reasoning (ANSR) program under award number FA8750-23-2-1016. The work has received partial funding from The Australian Research Council Discovery Project ARC DP2020102427.}
}
\begin{document}

\maketitle
\thispagestyle{empty}
\pagestyle{empty}

\begin{abstract}

We propose \MethodName{}, a semantic 3D Gaussian Splatting SLAM method featuring a novel hierarchical categorical representation, which enables accurate global 3D semantic mapping, scaling-up capability, and explicit semantic label prediction in the 3D world.
The parameter usage in semantic SLAM systems increases significantly with the growing complexity of the environment, making it particularly challenging and costly for scene understanding. 
To address this problem, we introduce a novel hierarchical representation that encodes semantic information in a compact form into 3D Gaussian Splatting, leveraging the capabilities of large language models (LLMs). 
We further introduce a novel semantic loss designed to optimize hierarchical semantic information through both inter-level and cross-level optimization. 
Furthermore, we enhance the whole SLAM system, resulting in improved tracking and mapping performance. Our \MethodName{} outperforms existing dense SLAM methods in both mapping and tracking accuracy, while achieving a 2x operation speed-up.
Additionally, it achieves on-par semantic rendering performance compared to existing methods while significantly reducing storage and training time requirements.
Rendering FPS impressively reaches 2,000 with semantic information and 3,000 without it.
Most notably, it showcases the capability of handling the complex real-world scene with more than 500 semantic classes, highlighting its valuable scaling-up capability. The open-source code is available at \href{https://github.com/LeeBY68/Hier-SLAM}{\textcolor{blue}{https://github.com/LeeBY68/Hier-SLAM}}.

\end{abstract}

\section{INTRODUCTION}

Visual Simultaneous Localization and Mapping (SLAM) is a critical technique for ego-motion estimation and scene perception, widely employed in multiple robotics tasks for drones \cite{heng2014autonomous}, self-driving cars \cite{lategahn2011visual}, as well as in applications such as Augmented Reality (AR) and Virtual Reality (VR) \cite{chekhlov2007ninja}. 
Semantic information, which provides high-level knowledge about the environment, is fundamental for comprehensive scene understanding and essential for intelligent robots to perform complex tasks. Recent advancements in image segmentation and map representations have significantly enhanced the performance of Semantic Visual SLAM~\cite{chang2021kimera, li2023dns}. 

\begin{figure}[t!]		
    \centering  
    \includegraphics[width=0.5\textwidth, trim=10mm 65mm 90mm 0mm, clip]{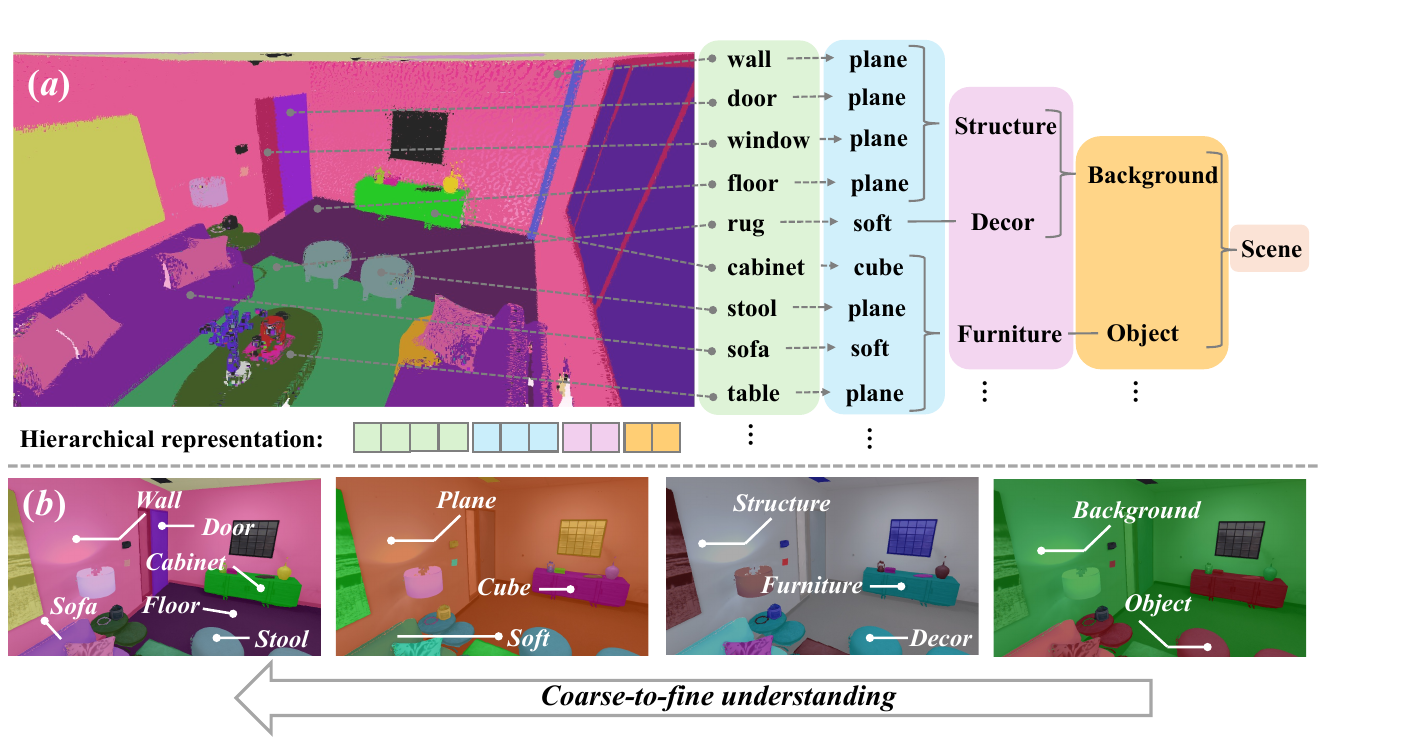} \\
    \vspace{-5pt}
    \caption{\textbf{(a). }The global 3D Gaussian map generated by \MethodName{} with learned semantic labels is shown on the left. The hierarchical structure of the semantic information is organized on the right, considering both \textbf{semantic} and \textbf{geometric} attributes (the second blue box). The proposed hierarchical categorical representation compresses semantic data, reducing both memory usage and training time of the semantic SLAM.
    \textbf{(b). }The rendered semantic map at different levels shows a coarse-to-fine understanding, beneficial for real-world scenarios with shifting perspectives from distant to close.}  
    \label{fig:hislam}  
    \vspace{-15pt}
\end{figure}

Recently, 3D Gaussian Splatting has emerged as a popular 3D world representation~\cite{kerbl20233d, Yu2024mipsplat,Wu20244dgs} due to its rapid rendering and optimization capabilities, attributed to the highly parallelized rasterization of 3D primitives. 
Specifically, 3D Gaussian Splatting effectively models the \textit{continuous} distributions of geometric parameters using Gaussian distribution. This capability not only enhances performance but also facilitates efficient optimization, which is especially advantageous for SLAM tasks. SLAM problem involves a complex optimization space, encompassing both camera poses and global map optimizations at the same time.
The adoption of 3D Gaussian Splatting has led to the development of several SLAM systems~\cite{keetha2023splatam, matsuki2024gaussian, yan2024gs, huang2024photo, yugay2023gaussian}, demonstrating promising performance in geometric understanding of unknown environments. 
However, the lack of semantic information in these approaches limits their ability to fully comprehend the global environment, restricting their potential in downstream tasks such as visual navigation, planning, and autonomous driving. 

Thus, it is highly desirable to extend the original 3D Gaussian Splatting with semantic capabilities while preserving its advantageous probabilistic representation. A straightforward approach would be to augment 3D points with a discrete semantic label and parameterize its distribution with a categorical discrete distribution, i.e., a flat Softmax embedding representation. 
However, 3D Gaussian Splatting is already a storage-intensive representation \cite{lu2024scaffold, chen2024hac}, requiring a large number of 3D primitives with multiple parameters to achieve realistic rendering. Adding semantic distribution parameters would result in significantly increased storage demands and processing time, growing linearly with the number of semantic classes. This makes it particularly impractical for complex scene understanding. 
Recent works formulate semantic classes using non-distributional approaches to handle this complexity.
The work \cite{li2024sgs} directly learns a 3-channel RGB visualization for semantic maps instead of the semantic label learning.
Another work \cite{zhu2024semgauss} uses a flat semantic representation with supervision from pre-trained foundation models to produces a 3D semantic embedding feature map. 

Unlike flat representations, semantic information naturally organises into a hierarchical structure of classes, as illustrated in \Fig{fig:hislam}.
This hierarchical relationship can be effectively represented as a tree structure, allowing for efficient encoding of extensive information with a relatively small number of nodes, i.e., a compact code. For instance, a binary tree with a depth of $10$ can cover $2^{10}$ classes, enabling the representation of $1,024$ classes using just symbolic 20 codes (i.e., $2 \times 10$, through 2-dimensional Softmax coding for each level). 

Building on this concept, we propose \MethodName{}, a Semantic Gaussian Splatting SLAM leveraging the hierarchical categorical representation for semantic information. 
Specifically, taking both semantic and geometric attributes into consideration, a well-designed tree is established with the help of Large Language Models (LLMs), which significantly reduces memory usage and training time, effectively compressing data while preserving its physical meaning. 
Additionally, we introduce a hierarchical loss for the proposed representation, incorporating both inter-level and cross-level optimizations. 
This strategy facilitates a coarse-to-fine understanding of scenes, which aligns well with real-world applications, particularly those involving observations from distant to nearby views. 
Furthermore, we enhance and refine the Gaussian SLAM to improve both performance and running speed. 

The main contributions of this paper include:

1) 
We propose a novel hierarchical representation that encodes semantic information by considering both geometric and semantic aspects, with assistance from LLMs. This tree coding effectively compacts the semantic information while preserving its physical hierarchical structure.

2)
We introduce a novel optimization loss for the semantic hierarchical representation, incorporating both inter-level and cross-level optimizations, ensuring comprehensive refinement across all levels of the hierarchical coding.

3) 
We conduct experiments on both synthetic and real-world datasets. The results demonstrate that our SLAM system outperforms existing methods in localization and mapping performance while achieving faster speeds. 
For semantic understanding, our method achieves on-par semantic rendering performance while significantly reducing storage and training time.
In complex real-world scenes, our approach, for the first time, demonstrates a valuable scaling-up capability, successfully handling more than 500 semantic classes—an important step toward the semantic understanding of complex environments.  

\begin{figure*}[t!]		
    \centering  
    \includegraphics[width=1.0\textwidth, trim=0mm 100mm 70mm 0mm, clip]{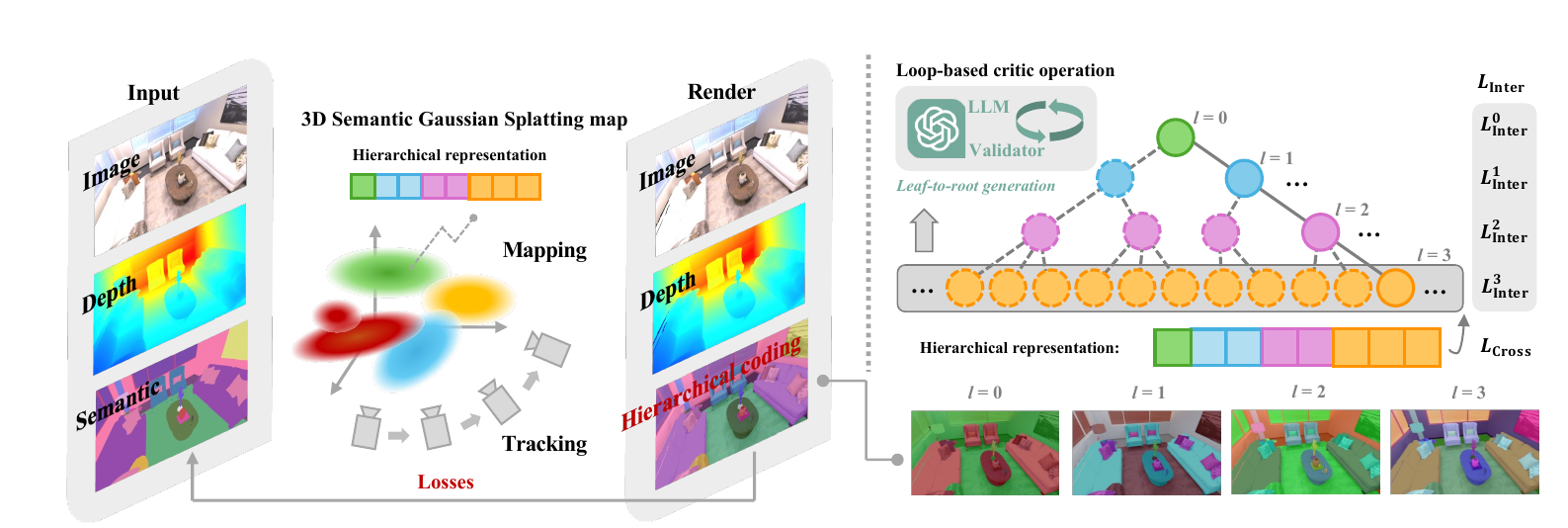} \\
    \caption{\textbf{Left:} Overview of the \MethodName{} pipeline. The global 3D Gaussian map is initialized with the first image. The system then alternates between \textit{Tracking} and \textit{Mapping} steps as new frames are processed (see Section III-C).
    \textbf{Top Right:} Hierarchical representation of semantic information. The Tree Generation process uses a Loop-based critic operation, including a LLM and a Validator, to create a tree coding from leaf-to-root. This tree is used to establish hierarchical coding for each Gaussian primitive (see Section III-A). Additionally, a novel loss combining Inter-level Loss $L_\text{Inter}$ and Cross-level Loss $L_\text{Cross}$ is proposed for hierarchical semantic optimization (see Section III-B). \textbf{Bottom Right:} An example of hierarchical semantic rendering.
    }  
    \label{fig:slam_pipeline}  
    \vspace{-15pt}
\end{figure*}

\section{RELATED WORK}

\paragraph{3D Gaussian Splatting SLAM}
    3D Gaussian Splatting has emerged as a promising 3D representation recently. 
    With the usage of 3D Gaussian Splatting, 
    SplaTAM \cite{keetha2023splatam} leverages silhouette guidance for pose estimation and map reconstruction in RGBD SLAM systems. MonoGS \cite{matsuki2024gaussian} implements both monocular and RGBD SLAM using 3D Gaussian Splatting. 
    3D Gaussian Splatting has demonstrated its strong capabilities across various Gaussian Splatting SLAM tasks \cite{yan2024gs, huang2024photo, yugay2023gaussian}. 
    However, integrating semantic understanding into SLAM tasks makes optimization particularly challenging, as it combines three high-dimensional optimization problems with different value ranges and convergence characteristics that to be optimized jointly.
    In this paper, we leverage hierarchical coding for semantic information and employ a suitable optimization strategy to ensure effective optimization across the hierarchical representation.

\paragraph{Neural Implicit Semantic SLAM}
    Semantic SLAM has been a longstanding research topic in the field of computer vision and robotics \cite{chang2021kimera, li2023textslam, rosinol2020kimera, li2020textslam, sualeh2019simultaneous}.
    Many works \cite{li2023dns, zhu2023sni, haghighi2023neural} have utilized the neural implicit representation for semantic mapping and localization tasks. 
    DNS-SLAM \cite{li2023dns} leverages 2D semantic priors combined with a coarse-to-fine geometry representation to integrate semantic information into the established map. 
    SNI-SLAM \cite{zhu2023sni} incorporates appearance, geometry, and semantic features into a collaborative feature space to enhance the robustness of the entire SLAM system.
    However, these methods are constrained by the limitations of neural implicit map representations, which is known to suffer from slow convergence, which leads to inefficiency and performance degradation when combined with semantic objectives~\cite{Hu2022CVPR, Liu2023ICCV}.
    In contrast, Gaussian Splatting offers advantages with its fast rendering performance and high-density reconstruction quality at the same time.

\paragraph{Gaussian Splatting Semantic SLAM}
    With the recent emergence of 3D Gaussian Splatting, SGS-SLAM \cite{li2024sgs} integrated additional RGB 3-channels to learn semantic visualization map, rather than true semantic understanding. SemGauss-SLAM \cite{zhu2024semgauss} employs a flat semantic representation, supervised by a large pre-trained foundation model. 
    However, these methods neglect the natural hierarchical characteristics of the real world. Furthermore, the reliance on large foundation models increases the complexity of the neural network and its computational demands, with performance heavily dependent on the embeddings from these pre-trained models. 
    In this paper, we introduce a simple yet effective hierarchical representation for semantic understanding, eliminating the dependency on foundation models, enabling a coarse-to-fine semantic understanding for the unknown environments.

\section{METHOD}

\subsection{Hierarchical representation} 
\textbf{Tree Parametrization.} 
We propose a hierarchical tree representation to encode semantic information, represented as $G = (V, E)$.
The node set $V = \cup_{l=0}^{L}\left\{v_{l}\right\}$ comprises all classes, where $\left\{v_{l}\right\}$ represents the set of nodes at the $l$-th level of the tree.
The edge set $E = \cup_{m=0}^{L-1}\left\{e_{m}\right\}$ captures the subordination relationships, encompassing both semantic attribution and geometric prior knowledge. Similarly we use the subscript $m$ to indicate the level of the tree. 
In this way, the $i$-th semantic class $g^{i}$, regarded as a single leaf node in the tree view, can be expressed hierarchically as:
\beq
g^{i} = \{ v_{l}^{i}, e_{m}^{i} \mid l = 0,1,...,L; \; m = 0,1,...,L-1 \},
\eeq
which corresponds to the root-to-leaf path: $ g^{i} = v_{0}^{i} \xrightarrow{e_{0}} v_{1}^{i} \xrightarrow{e_{1}} \cdots \xrightarrow{e_{L-2}} v_{L-1}^{i} \xrightarrow{e_{L-1}} v_{L}^{i}$.
Take a leaf node class \texttt{'Wall'} as an example, a 4-level tree coding can be as:
$\left\{v_{0}^\text{wall}:\texttt{Background}\right\} \rightarrow \left\{v_{1}^\text{wall}: \texttt{Structure}\right\} \rightarrow \left\{v_{2}^\text{wall}: \texttt{Plane}\right\} \rightarrow \left\{v_{3}^\text{wall}: \texttt{Wall}\right\}$.
Among these nodes, the relationships such as `include' and `possessing' are represented by the edge information \( e_{m}: \rightarrow \). 
In this way, any semantic concept can be coded in a progressive, hierarchical manner, incorporating both semantic and geometric perspectives. 
Moreover, the standard flat representation can be seen as a single-level tree coding from the hierarchical viewpoint.

\textbf{LLM-based Tree Generation.}
We utilize Large Language Models (LLMs), GPT-4o-mini~\cite{gpt4o_mini2024}, to generate the hierarchical tree representation due to its efficient performance. 
Specifically, a set of semantic class labels is provided to the LLMs, which iteratively clusters them into groups, regarding as the coarser-level classes. This process is repeated layer by layer from leaf to root, ultimately forming a complete hierarchical tree.
However, when dealing with a large number of semantic class labels in a complex environment, the results are often unsatisfactory because the LLM tends to cluster only a subset of the input classes, leaving out many classes and incorrectly including unseen classes in the hierarchy. 

To address this issue, we employ a loop-based critic operation, including an LLM followed by a validator. Specifically, during the clustering process from the $l$-level to the $(l-1)$-level, the $l$-level semantic classes $\left\{v_{l}\right\}$ are used as the prompt input to the LLM. 
The LLM then generates the clustering result $ \left\{ v_l' \right\} \rightarrow \left\{ v'_{l-1} \right\}$. 
It is important to note that the clustering result $\left\{v_{l}'\right\}$ may differ from the input $\left\{v_{l}\right\}$, as the LLM may introduce unseen classes or omit certain semantic labels.
By comparing the clustering result $\left\{v_{l}'\right\}$ and the prompt input $\left\{v_{l}\right\}$, the validator will identify three components: the successfully grouped nodes $\left\{ v_l' \right\}^{\text{success}}$, the unseen classes $\left\{ v_l' \right\}^{\text{unseen}}$, and the omitted semantic nodes $\left\{ v_l' \right\}^{\text{omitted}}$.
The successfully grouped nodes $\left\{ v_l' \right\}^{\text{success}}$ will be retained, while the unseen classes $\left\{ v_l' \right\}^{\text{unseen}}$ will be removed.
Next, the omitted nodes $\left\{ v_l' \right\}^{\text{omitted}}$ are used as the input prompt for the LLM to do the clustering in the subsequent iteration. 
At the same iteration, the clustering nodes $\left\{ v'_{l-1} \right\}$ generated by previous iteration are also provided to the LLM as a reference, suggesting that $\left\{ v_l' \right\}^{\text{omitted}}$ can either be clustered into the previously generated clusters or form new groups. 
This procedure loops until $\left\{ v_l' \right\}^{\text{omitted}} = \emptyset$, indicating that no classes are omitted.
In this way, we obtain all clustering results from the $l$-level to the $(l-1)$-level.
The proposed loop-based critic operation progresses from leaf to root, terminating when the LLM generates fewer than $\theta$ clusters. We set 
$\theta = 4$, ensuring that the number of nodes in the finest level remains small.
It is worth noting that the tree generation is performed offline before the SLAM operation.

\textbf{Tree Encoding.} 
For each 3D Gaussian primitive, its semantic embedding $\bm{h}$ is composed of the embedding $\bm{h}^{l}$ of each level:
\beq
\bm{h} = f(\bm{h}^{l}) \in \mathbb{R}^N, \quad \bm{h}^{l} \in \mathbb{R}^{n}, \quad l=0,1,...,L
\eeq
\noindent where we use $l$ to represents $l$-th level of the tree and $f$ stands for the concatenation operation.
As shown in \Fig{fig:slam_pipeline}, the overall dimension of the hierarchical embedding is the sum of the dimensions across all levels $N=\sum_{l=0}^{L} n$, where the dimension $n$ of each embedding $\bm{h}^{l}$ is equal to the maximum number of nodes at the $l$-th level.


\vspace{-5pt}
\subsection{Hierarchical loss}

To fully optimize the hierarchical semantic coding effectively, we propose the hierarchical loss as follows:
\beq
L_\text{{Semantic}} = \omega_1 L_\text{{Inter}} + \omega_2 L_\text{{Cross}}
\eeq
\noindent where $L_\text{Inter}$ and $L_\text{Cross}$ stands for the Inter-level loss and Cross-level loss respectively. We use $\omega_1$ and $\omega_2$ to balance the weights between each loss.
The Inter-level loss $L_\text{Inter}$ is employed within each level:
\beq
L_\text{{Inter}} = \sum_{l=0}^{L} L_\text{{ce}}(\text{softmax}(\bm{h}^{l}), \mathcal{P}^{l})
\eeq
\noindent where $L_\text{{ce}}$ represents the cross-entropy loss, and $\mathcal{P}^{l}$ stands for the semantic ground truth for the $l$-th level. 
In contrast, the Cross-level loss is computed based on the entire hierarchical coding. First, a linear layer $F$ shared between all Gaussian primitives is used to transform the hierarchical embeddings into flat coding. Following is a $\text{softmax}(F(\bm{h}))$ operation to convert the embeddings into probabilities. The Cross-level loss \( L_{\text{Cross}} \) is then defined as follows:
\beq
L_{\text{Cross}} = L_{\text{ce}}\left( \text{softmax}(F(\bm{h})), \mathcal{P} \right)
\eeq
\noindent where $\mathcal{P}$ denotes semantic ground truth in flat representation.

\subsection{Gaussian Splatting Semantic Mapping and Tracking}

The pipeline of our \MethodName{} is illustrated in \Fig{fig:slam_pipeline}. We will detail the submodules in this subsection.

\textbf{Semantic 3D Gaussian representation.} 
We adopt Gaussian primitives with hierarchical semantic embedding for the scene representation.
Each semantic Gaussian is represented as the combination of color $\bm{c}$, the center position $\bm{\mu}$, the radius $r$, the opacity $o$, and its semantic embedding $\bm{h}$.  
And the influence of each Gaussian according to the standard Gaussian equation is 
$G = o \: \exp\left(-\frac{||\bm{X}-\bm{\mu}||^2}{2r^2}\right)$,
\noindent where $\bm{X}$ stands for the 3D point. 

Following \cite{kerbl20233d}, each semantic 3D Gaussian primitive is projected to the 2D image space using the tile-based differentiable $\alpha$-compositing rendering. 
The semantic map is rasterized as follows:
\begin{equation}
H = \sum_{i=1}^{n} \bm{h}_{i} G_{i}(\bm{X}) T_{i} \quad \text{with} \quad T_{i} = \prod_{j=1}^{i-1} (1 - G_{j}(\bm{X}))
\end{equation}
The rendered color image $C$, depth image $D$, and the silhouette image $S$ are defined as follows:
\begin{align}
C = \sum_{i=1}^{n} \bm{c}_{i} G_{i}(\bm{X}) T_{i}, \;
D = \sum_{i=1}^{n} \bm{d}_{i} G_{i}(\bm{X}) T_{i}, \;
S = \sum_{i=1}^{n} G_{i}(\bm{X}) T_{i}
\end{align}
In contrast to previous work \cite{keetha2023splatam}, which employs separate forward and backward rendering modules for different parameters, we adopt unified forward and backward modules that handle all parameters, including semantic, color, depth, and silhouette images, significantly improving the overall efficiency of the SLAM system.

\textbf{Tracking.} 
The tracking step aims to estimate each frame's pose. We adopt constant velocity model to initialize the pose of every incoming frame, following a pose optimization while fixing the global map, using the rendering color and depth losses:
\beq
L_\text{{Track}} = M \left( w_{1} L_\text{{Depth}} + w_{2} L_\text{{Color}} \right)
\eeq
\noindent where $L_\text{Depth}$ and $L_\text{Color}$ stands for the L1-loss for the rendered depth and color information.
We use weights $ w_{1}$ and $w_{2}$ to balance the two losses and the optimization is only performed on the silhouette-visible image $M = (S > \delta)$.

\textbf{Mapping.} 
The global map information, including the semantic information, is optimized in the mapping procedure with fixed camera poses. 
The optimization losses include the depth, color, and the semantic losses:
\beq
L_\text{{Map}} = w_{3}  M  L_\text{{Depth}} + w_{4} L_\text{{Color}}' + w_{5} L_\text{{Semantic}} 
\eeq 
\noindent where $L_\text{{Semantic}}$ is the proposed semantic loss introduced in Section III-B, and $L_\text{{Color}}'$ is the weighted sum of SSIM color loss and L1-Loss. And we use $w_{3}$, $w_{4}$, and $w_{5}$ for balancing different terms.

\vspace{-10pt}
\section{Experiments}
\vspace{-5pt}

\begin{figure*}[t!]		
    \centering  
    \includegraphics[width=0.85\textwidth, trim=40mm 35mm 60mm 20mm, clip]{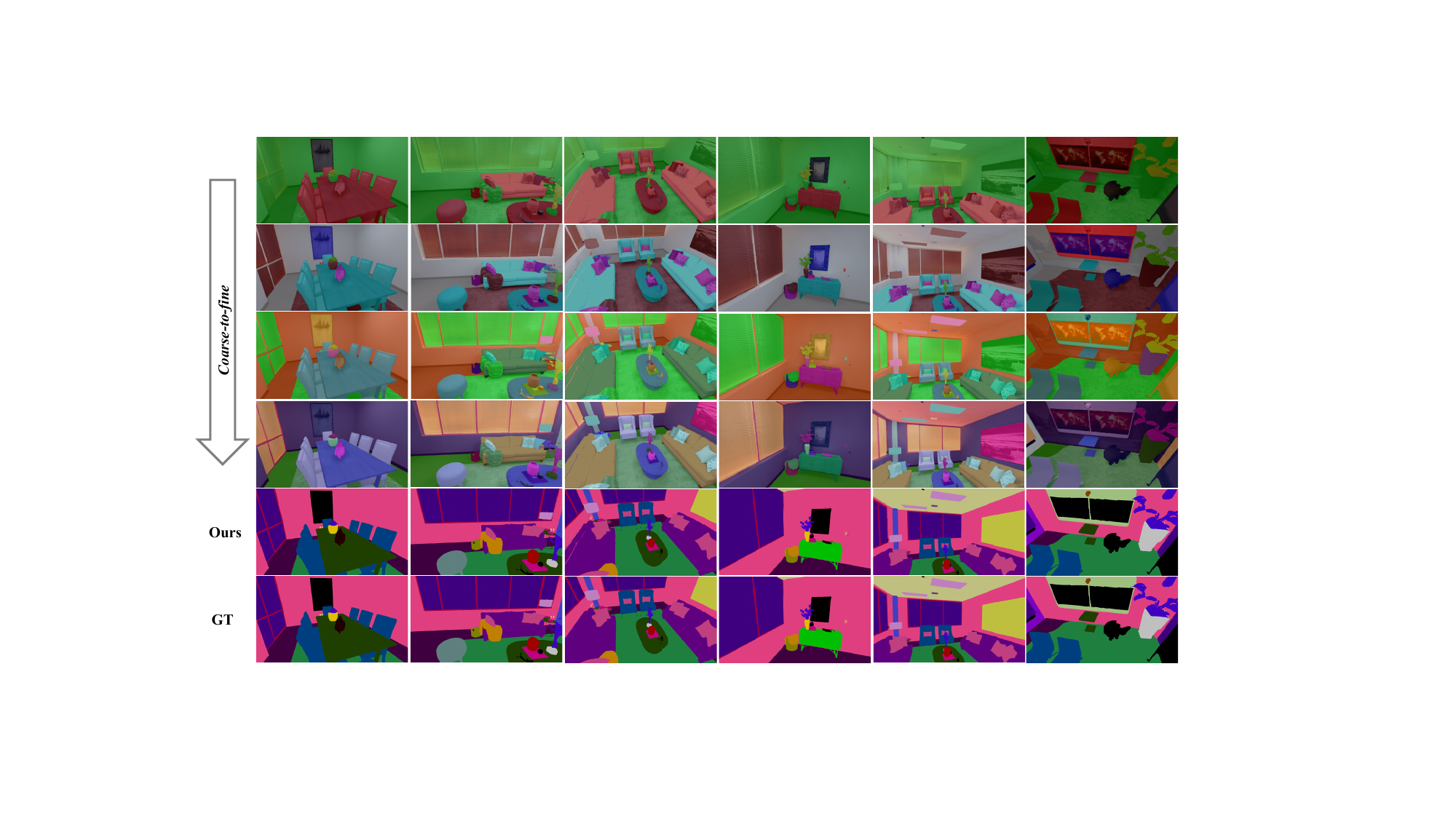} \\
    \vspace{-5pt}
    \caption{Visualization of our semantic rendering performance on the Replica \cite{straub2019replica} dataset. \textbf{The first four rows} demonstrate rendered semantic segmentation in a coarse-to-fine manner. \textbf{The fifth row} exhibits the finest semantic rendering, equivalent to the flat representation with $102$ original semantic classes from the Replica dataset. \textbf{The last row} visualizes the semantic ground truth for comparison.}  
    \label{fig:exp_sem_render} 
    \vspace{-15pt}
\end{figure*}

\subsection{Experiment settings}
\vspace{-5pt}

The experiments are conducted on both synthetic and real-world datasets, including 6 scenes from ScanNet \cite{Dai2017scannet} and 8 sequences from Replica \cite{straub2019replica}.
Following the evaluation metrics used in previous SLAM works \cite{keetha2023splatam, zhu2022nice}, we leverage ATE RMSE (cm) to assess SLAM tracking accuracy. 
For mapping performance, we use Depth L1 (cm) to evaluate accuracy. 
To assess image rendering quality, we adopt PSNR (dB), SSIM, and LPIPS metrics.
Similar to previous methods \cite{zhu2024semgauss, li2023dns,zhu2023sni}, due to the lack of direct metrics for evaluating 3D semantic understanding in 3D Gaussian Splatting representations, we rely on 2D semantic segmentation performance, measured by mIoU (mean Intersection over Union across all classes), to reflect global semantic information. 
To demonstrate the improved efficiency, we also measure the running time of the proposed SLAM method.
We compare our method against state-of-the-art dense visual SLAM approaches, including both NeRF-based and 3D Gaussian SLAM methods, to highlight its effectiveness. Additionally, we include state-of-the-art semantic SLAM techniques, covering both NeRF-based and Gaussian-based methods, to showcase our hierarchical semantic understanding and scaling-up capability.
The experiments are conducted in the Nvidia L40S GPU.
For experimental settings, the semantic embedding of each Gaussian primitive is initialized randomly. We set semantic optimization loss weights $\omega_1$ and $\omega_2$ to$1.0$ and $0.0$, respectively, for the first $\eta$ iterations, where $\eta$ is set to $15$. Afterwards, $\omega_1$ and $\omega_2$ are adjusted to $1.0$ and $5.0$, respectively. This means that we first use the Inter-level loss to initialize the hierarchical coding, followed by incorporating the Cross-level loss to refine the embedding. 
For tracking loss, we set $\delta=0.99$, $w_{1}=1.0$, $w_{2}=0.5$.
For mapping, we set $w_{3}=1.0$, $w_{4}=0.5$, $w_{5}=0.2$, respectively.

\vspace{-5pt}
\subsection{SLAM Performance}
\vspace{-3pt}

\textbf{Tracking Accuracy.}
We present the tracking performance on the Replica \cite{straub2019replica} and ScanNet \cite{Dai2017scannet} datasets in \Tab{tab:exp_pose_replica} and \Tab{tab:exp_pose_scannet}, respectively. 
On the Replica dataset, our proposed method surpasses all current approaches.
For the ScanNet dataset, the performance of all methods is lower than on the synthetic dataset due to the noisy, sparse depth sensor input and the limited color image quality caused by motion blur. We evaluate all six sequences, showing that our method performs comparably to state-of-the-art methods \cite{keetha2023splatam,zhu2022nice}.

\begin{table}[!t]
    \centering
    \caption{Localization performance ATE RMSE (cm) on the Replica dataset. Best results are highlighted as \colorbox{green!30}{\textbf{FIRST}}, \colorbox{yellow!30}{SECOND}.} 
    \vspace{-10pt}
    \renewcommand{\arraystretch}{1.1} 
    \setlength{\tabcolsep}{2pt}
    \begin{tabular}{lccccccccc}
        \hline
        \rowcolor{white}
        \toprule
        \textbf{Methods} & \textbf{Avg.} & \textbf{R0} & \textbf{R1} & \textbf{R2} & \textbf{Of0} & \textbf{Of1} & \textbf{Of2} & \textbf{Of3} & \textbf{Of4} \\
        \hline
        iMap \cite{sucar2021imap} & 4.15 & 6.33 & 3.46 & 2.65 & 3.31 & 1.42 & 7.17 & 6.32 & 2.55 \\
        NICE-SLAM \cite{zhu2022nice} & 1.07 & 0.97 & 1.31 & 1.07 & 0.88 & 1.00 & 1.06 & 1.10 & 1.13 \\
        Vox-Fusion \cite{yang2022vox} & 3.09 & 1.37 & 4.70 & 1.47 & 8.48 & 2.04 & 2.58 & 1.11 & 2.94 \\
        co-SLAM \cite{wang2023co} & 1.06 & 0.72 & 0.85 & 1.02 & 0.69 & 0.56 & 2.12 & 1.62 & 0.87 \\
        ESLAM \cite{johari2023eslam} & 0.63 & 0.71 & 0.70 & 0.52 & 0.57 & 0.55 & 0.58 & 0.72 & 0.63 \\
        Point-SLAM \cite{sandstrom2023point} & 0.52 & 0.61 & 0.41 & 0.37 & \cellcolor{yellow!30}0.38 & 0.48 & 0.54 & 0.69 & 0.72 \\
        MonoGS \cite{matsuki2024gaussian} & 0.79 & 0.47 & 0.43 & 0.31 & 0.70 & 0.57 & 0.31 & \cellcolor{green!30}0.31 & 3.2 \\  
        SplaTAM \cite{keetha2023splatam} & \cellcolor{yellow!30}0.36 & \cellcolor{yellow!30}0.31 & \cellcolor{green!30}0.40 & \cellcolor{yellow!30}0.29 & 0.47 & \cellcolor{yellow!30}0.27 & \cellcolor{green!30}0.29 & \cellcolor{yellow!30}0.32 & \cellcolor{yellow!30}0.55 \\
        \hdashline
        \textbf{\MethodName{} (Ours*)} & \cellcolor{green!30}\textbf{0.32} & \cellcolor{green!30}0.24 & \cellcolor{yellow!30}0.44 & \cellcolor{green!30}0.25 & \cellcolor{green!30}0.28 & \cellcolor{green!30}0.17 & \cellcolor{green!30}0.29 & 0.37 & \cellcolor{green!30}0.49 \\ 
        \hline
        SNI-SLAM \cite{zhu2023sni} & 0.46 & 0.50 & 0.55 & 0.45 & 0.35 & 0.41 & 0.33 & 0.62 & \cellcolor{yellow!30}0.50 \\
        DNS SLAM \cite{li2023dns} & 0.45 & 0.49 & 0.46 & 0.38 & \cellcolor{yellow!30}0.34 & 0.35 & 0.39 & 0.62 & 0.60 \\
        SemGauss-SLAM \cite{zhu2024semgauss} & \cellcolor{yellow!30} 0.33 & \cellcolor{yellow!30} 0.26 & \cellcolor{green!30}0.42 & \cellcolor{yellow!30}0.27 & \cellcolor{yellow!30}0.34 & \cellcolor{yellow!30}0.17 & \cellcolor{yellow!30}0.32 & \cellcolor{green!30}0.36 & \cellcolor{green!30}0.49 \\
        \hdashline
        \textbf{\MethodName{} (Ours)} &  \cellcolor{green!30}\textbf{0.33} &  \cellcolor{green!30}0.21 & \cellcolor{yellow!30}0.49 & \cellcolor{green!30}0.24 & \cellcolor{green!30}0.29 & \cellcolor{green!30}0.16 & \cellcolor{green!30}0.31 & \cellcolor{yellow!30}0.37 & 0.53 \\ 
         \toprule
    \end{tabular}
    \label{tab:exp_pose_replica}
    \begin{tablenotes} 
            \footnotesize
            \itshape
            \raggedright 
            \item \textbf{Ours*} represents our proposed system without semantic information.
            \vspace{-20pt}
        \end{tablenotes}
\end{table}

\begin{table}[hb]
\vspace{-10pt}
    \centering
    \setlength{\tabcolsep}{2pt}
    \caption{Localization performance ATE RMSE (cm) on the Scannet dataset. Best results are highlighted as \colorbox{green!30}{first}, \colorbox{yellow!30}{second}, \colorbox{red!30}{third}.}
    \vspace{-10pt}
    \label{tab:exp_pose_scannet}
    \begin{tabular}{lccccccc}
        \toprule
        \textbf{Methods} & \textbf{Avg.} & \textbf{0000} & \textbf{0059} & \textbf{0106} & \textbf{0169} & \textbf{0181} & \textbf{0207}  \\
        \midrule
        NICE-SLAM \cite{zhu2022nice} & \cellcolor{green!30}10.70 & \cellcolor{red!30}12.00 & 14.00 & \cellcolor{green!30}7.90 & \cellcolor{yellow!30}10.90 & 13.40 & \cellcolor{green!30}6.20  \\
        Vox-Fusion \cite{yang2022vox} & 26.90 & 68.84 & 24.18 & \cellcolor{yellow!30}8.41 & 27.28 & 23.30 & 9.41  \\
        Point-SLAM \cite{sandstrom2023point} & 12.19 & \cellcolor{green!30}10.24 & \cellcolor{green!30}7.81 &\cellcolor{red!30}8.65 & 22.16 & 14.77 & 9.54  \\
        SplaTAM \cite{keetha2023splatam} & 11.88 & 12.83 & 10.10 &17.72 & 12.08 & \cellcolor{red!30}11.10 & \cellcolor{red!30}7.46  \\
        SemGauss-SLAM \cite{zhu2024semgauss} & -- & 11.87 & \cellcolor{yellow!30}7.97 & -- & \cellcolor{green!30}8.70 & \cellcolor{green!30}9.78 & 8.97  \\
        \hdashline
        \MethodName{} (Ours*) & \colorbox{red!30}{11.80} & 12.83 & \colorbox{red!30}{9.57} & 17.54 & \cellcolor{red!30}11.54 & 11.78 & 7.55 \\
        \MethodName{} (Ours) & \cellcolor{yellow!30}11.36 & \cellcolor{yellow!30}11.45 & 9.61 & 17.80 & 11.93 & \cellcolor{yellow!30}10.04 & \cellcolor{yellow!30}7.32 \\
        \bottomrule 
    \end{tabular}
    \begin{tablenotes} 
            \footnotesize
            \itshape
            \raggedright
            \item \textbf{Ours*} represents our proposed system without semantic information.
            \vspace{-10pt}
        \end{tablenotes}
\end{table}

\textbf{Mapping Performance.}
In \Tab{tab:exp_depth}, we evaluate the mapping performance using the L1 depth loss in Replica \cite{straub2019replica}. The results show that our method surpasses all existing approaches, demonstrating superior mapping capabilities.

\textbf{Rendering Quality.}
Similar to Point-SLAM \cite{sandstrom2023point} and NICE-SLAM \cite{zhu2022nice}, we evaluate rendering quality on input views from 8 sequences of the Replica dataset \cite{straub2019replica}. The evaluation uses average PSNR, SSIM, and LPIPS metrics.
Our methods achieves superior performance (\MethodName{}: $\text{PSNR}\uparrow: 35.70, \, \text{SSIM}\uparrow: 0.980, \, \text{LPIPS}\downarrow: 0.067$) compared to the state-of-the-art approaches, where the best performances being: (SpltaTAM \cite{keetha2023splatam}: $\text{PSNR}\uparrow: 34.11, \, \text{SSIM}\uparrow: 0.968, \, \text{LPIPS}\downarrow: 0.102$), and (SemGauss-SLAM \cite{zhu2024semgauss}: $\text{PSNR}\uparrow: 35.03, \, \text{SSIM}\uparrow: 0.982, \, \text{LPIPS}\downarrow: 0.062$).
Detail performances are provided in the Appendix.

\begin{table}[!t]
    \centering
    \caption{Reconstruction metric Depth L1 (cm) comparison on Replica. Best results are highlighted as \colorbox{green!30}{first}, \colorbox{yellow!30}{second}.}
    \vspace{-10pt}
    \renewcommand{\arraystretch}{1.1} 
    \setlength{\tabcolsep}{1.5pt}     
    \begin{tabular}{lccccccccc}
        \hline
        \toprule
        \textbf{Methods} & \textbf{Avg.} & \textbf{R0} & \textbf{R1} & \textbf{R2} & \textbf{Of0} & \textbf{Of1} & \textbf{Of2} & \textbf{Of3} & \textbf{Of4} \\
        \hline
        NICE-SLAM \cite{zhu2022nice} & 2.97 & 1.81 & 1.44 & 2.04 & 1.39 & 1.76 & 8.33 & 4.99 & 2.01 \\
        Vox-Fusion \cite{yang2022vox} & 2.46 & 1.09 & 1.90 & 2.21 & 2.32 & 3.40 & 4.19 & 2.96 & 1.61 \\
        Co-SLAM \cite{wang2023co} & 1.51 & 1.05 & 0.85 & 2.37 & 1.24 & 1.48 & 1.86 & 1.66 & 1.54 \\
        ESLAM \cite{johari2023eslam} & 0.95 & 0.73 & 0.74 & 1.26 & 0.71 & 1.02 & 0.93 & 1.03 & 1.18 \\
        SNI-SLAM \cite{zhu2023sni} & 0.77 & \cellcolor{yellow!30}0.55 & 0.58 & 0.87 & 0.55 & 0.97 & 0.89 & \cellcolor{green!30}0.75 & 0.97 \\
        SemGauss-SLAM \cite{zhu2024semgauss} & \cellcolor{yellow!30}0.50 & \cellcolor{green!30}0.54 & \cellcolor{yellow!30}0.46 & \cellcolor{yellow!30}0.43 & \cellcolor{green!30}0.29 & \cellcolor{yellow!30}0.22 & \cellcolor{green!30}0.51 & 0.98 & \cellcolor{green!30}0.56 \\
        \hdashline
        \MethodName{} (Ours) & \cellcolor{green!30}0.49 & 0.58 & \cellcolor{green!30}0.40 & \cellcolor{green!30}0.40 & \cellcolor{green!30}0.29 & \cellcolor{green!30}0.19 & \cellcolor{green!30}0.51 & \cellcolor{yellow!30}0.95 & \cellcolor{yellow!30}0.57 \\
        \toprule
    \end{tabular}
    \label{tab:exp_depth}
    \vspace{-13pt}
\end{table}

\textbf{Running time.}
Running times for all methods are shown in \Tab{tab:runtime}. Compared to state-of-the-art dense visual SLAM approaches, our method (Ours*) achieves up to 2.4× faster tracking and 2.2× faster mapping than the SOTA performance \cite{keetha2023splatam}. 
When incorporating semantic information, our method remains efficient, leveraging hierarchical semantic coding to achieve nearly 3× faster tracking and 1.2× faster mapping compared with the semantic SLAM with flat semantic coding.
Notably, our \MethodName{} achieves \textbf{a rendering speed of 2000 FPS}. For \MethodName{} without semantic information, \textbf{the rendering speed increases to 3000 FPS}.  

\begin{table}
\centering
\setlength{\tabcolsep}{2pt}
\caption{Runtime on Replica/R0. Best results are highlighted as \textbf{first}.}
\vspace{-5pt}
\label{tab:runtime}
\begin{tabular}{lcccc}
\hline
\toprule
\multirow{2}{*}{Methods} & Tracking & Mapping & Tracking & Mapping \\
 & /Iteration (ms) & /Iteration (ms) & /Frame (s) & /Frame (s) \\
\hline
NICE-SLAM \cite{zhu2022nice} & 122.42 & 104.25 & 1.22 & 6.26 \\
SplaTAM \cite{keetha2023splatam} & 44.27 & 50.07 & 1.77 & 3.00 \\
\MethodName{} (Ours*) & \textbf{18.71} & \textbf{22.93} & \textbf{0.75} & \textbf{1.38} \\
\MethodName{} (Ours) & 46.90 & 148.66 & 1.88 & 8.92 \\ 
\hline
\MethodName{} (Ours) & \textbf{61.23} & \textbf{170.30} & \textbf{2.45} & \textbf{10.22} \\
\MethodName{} (Ours**) & 168.94 & 204.25 & 6.75 & 12.26 \\
\toprule
\end{tabular}
\begin{tablenotes} 
            \footnotesize
            \itshape
            \raggedright 
            \item \textbf{First \& Second block} results are from NVIDIA GeForce RTX 4090 and NVIDIA L40S, respectively.
            \item \textbf{Ours*} represents our proposed system without semantic information.
            \item \textbf{Ours**} represents our proposed system using flat semantic encoding.
            \vspace{-5pt}
        \end{tablenotes}
\vspace{-14pt}
\end{table}

\vspace{-5pt}
\subsection{Hierarchical semantic understanding}
\vspace{-2pt}

We conduct semantic understanding experiments in synthetic dataset Replica \cite{straub2019replica} to demonstrate the comprehensive performance of our proposed method. 
Replica \cite{straub2019replica} is a synthetic indoor dataset comprising a total of 102 semantic classes with high-quality semantic ground truth.

We establish a five-level tree to encode these original classes hierarchically. The semantic rendering performance is illustrated in \Fig{fig:exp_sem_render}, where the first five rows show the progression from level-0 to level-4, moving from coarse to fine understanding.
The coarsest semantic rendering, i.e., level-0 which shown in the first row, includes segmentation covering 4 broad classes: \textit{Background}, \textit{Object}, \textit{Other}, and \textit{Void}. In contrast, the finest level encompasses all 102 original semantic classes.
For example, the hierarchical understanding of the class \textit{'Stool'} progresses from $\textit{Object} \rightarrow \textit{Furniture} \rightarrow \textit{Plane} \rightarrow \textit{Chair} \rightarrow \textit{Stool}$, as depicted in the second column.
From \Fig{fig:exp_sem_render}, we observe that our method achieves precise semantic rendering at each level, providing a comprehensive coarse-to-fine semantic understanding for overall scenes.

Similar to previous methods \cite{zhu2024semgauss, li2023dns, zhu2023sni}, we present our quantitative results, evaluated in mIoU ($\%$) across all original semantic classes (102 classes) in \Tab{tab:replica_semantic_segmentation}, where the rendered semantic map is compared against the semantic ground truth. Additionally, we also report mIoU evaluated on a subset of semantic classes in the second block of \Tab{tab:replica_semantic_segmentation}.  
To demonstrate the efficiency of our proposed method, we report the storage usage (MB) and runtime in the last part of \Tab{tab:replica_semantic_segmentation} and \Tab{tab:runtime}, respectively.
From \Tab{tab:replica_semantic_segmentation}, our flat coding version achieves the best performance among all methods, attaining an mIoU of $90.35\%$ on all 102 semantic classes, at the cost of large storage usage.  
In contrast, the hierarchical representation achieves a competitive mIoU while requiring only $910.50$ MB of storage, which is $66\%$ less than the flat version.  
Since works \cite{zhu2023sni, zhu2024semgauss, li2024sgs} report mIoU only on a subset of classes, making direct comparison unfair, we report the same evaluation procedure as \cite{li2024sgs}. Our method achieves an mIoU of $95.58\%$ with a storage usage of $910.50$ MB, demonstrating superior semantic rendering performance compared to state-of-the-art methods \cite{zhu2023sni, li2024sgs} while maintaining efficient storage usage.   
Meanwhile, \cite{zhu2024semgauss} benefits significantly from a large foundation model pre-trained on much larger and more diverse datasets, making the comparison less fair.  
In terms of training time, \Tab{tab:runtime} shows that our proposed method requires only $36\%$ of the time for frame tracking and $83\%$ for frame mapping compared to the flat version.  
Overall, our method achieves performance on par with SOTA semantic SLAM while significantly reducing both storage requirements and training time, benefiting from the proposed hierarchical semantic representation.

\vspace{-10pt}
\begin{table}[!thb]
\centering
\setlength{\tabcolsep}{2pt}
\caption{Semantic performance mIoU (\%) and Parameter usage (MB) on Replica. Results are highlighted as \colorbox{green!30}{first}, \colorbox{yellow!30}{second}.}
\label{tab:replica_semantic_segmentation}
\vspace{-5pt}
\begin{tabular}{c|l|ccccccc}
\hline
& \textbf{Methods} & \textbf{Avg.} & \textbf{R0} & \textbf{R1} & \textbf{R2} & \textbf{Of0} \\ 
\hline
\multirow{4}{*}{\makecell{\textbf{mIoU (\%)} \\ total 102 classes}}
& NIDS-SLAM \cite{haghighi2023neural} & 82.37 & 82.45 & 84.08 & 76.99 & 85.94 \\ 
& DNS-SLAM \cite{li2023dns} & \secondcolor{84.77} & \secondcolor{88.32} & \secondcolor{84.90} & \secondcolor{81.20} & \secondcolor{84.66} \\ 
& \MethodName{} (Ours**) & \bestcolornob{90.35} & \bestcolornob{91.21} & \bestcolornob{90.62} & \bestcolornob{89.11} & \bestcolornob{90.45} \\ 
& \MethodName{} (Ours) & 76.44 & 76.62 & 78.31 & 80.39 & 70.43 \\  
\hdashline
\multirow{4}{*}{\makecell{\textbf{mIoU (\%)} \\ subset classes}}    & SNI-SLAM \cite{zhu2023sni} & 87.41 & 88.42 & 87.43 & 86.16 & 87.63 \\ 
 & SemGauss-SLAM \cite{zhu2024semgauss} & \bestcolornob{96.34} & \bestcolornob{96.30} & \bestcolornob{95.82} & \bestcolornob{96.51} & \bestcolornob{96.72} \\
 & SGS-SLAM \cite{li2024sgs} & 92.72 & 92.95 & 92.91 & 92.10 & 92.90 \\
 & \MethodName{} (Ours$^{\dag}$) & \secondcolor{95.58} & \secondcolor{95.25} &  \secondcolor{95.81} & \secondcolor{95.73}  & \secondcolor{95.52} \\   
\hline
\multirow{2}{*}{\textbf{Param (MB)}} 
& \MethodName{} (Ours**) & 2662.25 & 2355 & 3072 & 2560 & 2662\\  
& \MethodName{} (Ours) & \cellcolor{green!30}910.50 & \cellcolor{green!30}793 & \cellcolor{green!30}1126 & \cellcolor{green!30}843 & \cellcolor{green!30}880 \\ 
\hline
\end{tabular}
\begin{tablenotes} 
            \footnotesize
            \itshape
            \raggedright
            \item \textbf{Ours**} represents our proposed system using flat semantic encoding.
            \item \textbf{Ours$^{\dag}$} represents our method with a hierarchical representation, evaluated on a subset of semantic classes, consistent with \cite{li2024sgs}.
            \item \textbf{First \& Second block} mIou (\%) results are evaluated over a total of 102 semantic classes and a subset of classes, respectively.
            \vspace{-10pt}
        \end{tablenotes}
\end{table}

\begin{figure}[t!]		
    \centering  
    \includegraphics[width=0.5\textwidth, trim=0mm 70mm 0mm 0mm, clip]{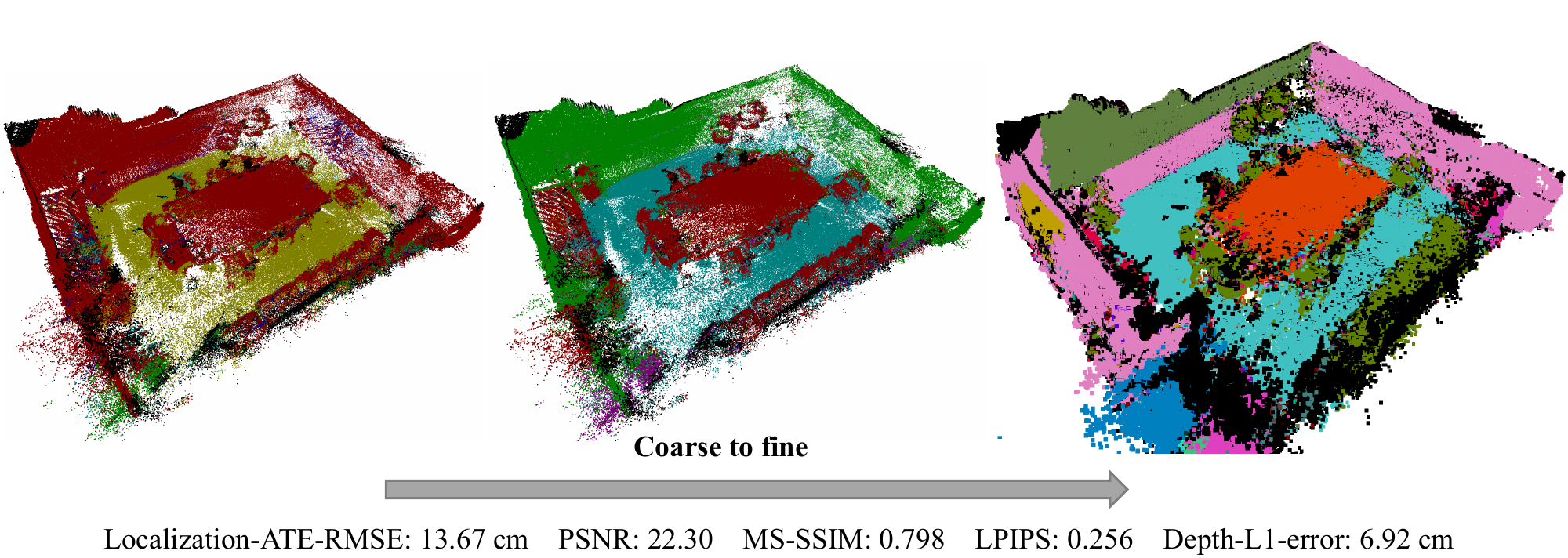} \\
    \vspace{-5pt}
    \caption{Visualization of the established semantic 3D map across multiple levels, demonstrating a coarse-to-fine semantic understanding of the complex scene. The bottom of the figure displays localization, mapping, and rendering performance, providing a comprehensive overview.}  
    \label{fig:scannet_scaleup}  
    \vspace{-15pt}
\end{figure}

\vspace{3pt}
\subsection{Scaling up capability}

To demonstrate the scaling-up capability, we apply our proposed method to the real-world complex dataset, ScanNet \cite{Dai2017scannet}, which covers up to 550 unique semantic classes. Unlike Replica \cite{straub2019replica}, where the semantic ground truth is synthesized from a global world model and can be considered ideal, the semantic annotations in ScanNet are significantly noisier. Additionally, the dataset features noisy depth sensor inputs and blurred color images, making semantic understanding particularly challenging in this scenes.
Using the flat semantic representation cannot even run successfully due to storage limitations. In contrast, we establish the hierarchical tree with the assistance of LLMs, which guide the compaction of the coding from the original 550 semantic classes to 72 semantic codings, resulting in over 7 times reduction in coding usage. As visualized in \Fig{fig:scannet_scaleup}, our estimated 3D global semantic map at different levels demonstrates a coarse-to-fine semantic understanding, showcasing our method's scaling-up capability in handling this complex scene.

\vspace{-5pt}
\section{CONCLUSIONS}
\vspace{-5pt}
We present \MethodName{}, a novel semantic 3D Gaussian Splatting SLAM method with a hierarchical categorical representation, which can generate explicit global 3D semantic label map with scaling-up capability. 
Specifically, we introduce a compact hierarchical representation for semantic encoding, integrating it into 3D Gaussian Splatting with LLMs assistance.
We further propose a novel semantic loss for optimizing hierarchical semantic information across inter- and cross-levels.
Our refined SLAM system achieves superior or on-par performance with existing dense SLAM methods in tracking, mapping, and semantic understanding while running faster and significantly reducing storage. \MethodName{} delivers exceptional rendering at up to 2,000/3,000 FPS (with/without semantics) and effectively handles complex real-world scenes, demonstrating strong scalability.









\bibliographystyle{unsrt}
\bibliography{references}

\clearpage
\newpage
\setcounter{section}{0}
\begin{center}
    \textbf{\large APPENDIX}
\end{center}

\begin{strip}
    \centering
    \captionof{table}{Rendering performance PSNR, SSIM, LPIPS on Replica. Best results are highlighted as \colorbox{green!30}{first}, \colorbox{yellow!30}{second}.}
    \label{tab:exp_render_replica}
    \begin{tabular}{llcccccccccccc}
        \toprule
        \textbf{Methods} & \textbf{Metrics} & \textbf{Avg.} & \textbf{room0} & \textbf{room1} & \textbf{room2} & \textbf{office0} & \textbf{office1} & \textbf{office2} & \textbf{office3} & \textbf{office4} \\
        \midrule
        \multicolumn{11}{c}{\textbf{Visual SLAM}} \\
        \multirow{3}{*}{NICE-SLAM \cite{zhu2022nice}} 
        & PSNR $\uparrow$ & 24.42 & 22.12 & 22.47 & 24.52 & 29.07 & 30.34 & 19.66 & 22.23 & 24.94 \\
        & SSIM $\uparrow$ & 0.809 & 0.689 & 0.757 & 0.814 & 0.874 & 0.886 & 0.797 & 0.801 & 0.856 \\
        & LPIPS $\downarrow$ & 0.233 & 0.330 & 0.271 & 0.208 & 0.229 & 0.181 & 0.235 & 0.209 & 0.198  \\
        \hline
        \multirow{3}{*}{Vox-Fusion \cite{yang2022vox}} 
        & PSNR $\uparrow$ & 24.41 & 22.39 & 22.36 & 23.92 & 27.79 & 29.83 & 20.33 & 23.47 & 25.21 \\
        & SSIM $\uparrow$ & 0.801 & 0.683 & 0.751 & 0.798 & 0.857 & 0.876 & 0.794 & 0.803 & 0.847 \\
        & LPIPS $\downarrow$ & 0.236 & 0.303 & 0.269 & 0.234 & 0.241 & 0.184 & 0.243 & 0.213 & 0.199 \\
        \hline
        \multirow{3}{*}{Co-SLAM \cite{wang2023co}} 
        & PSNR $\uparrow$ & 30.24 & 27.27 & 28.45 & 29.06 & 34.14 & 34.87 & 28.43 & 28.76 & 30.91  \\
        & SSIM $\uparrow$ & 0.939 & 0.910 & 0.909 & 0.932 & 0.961 & 0.969 & 0.938 & 0.941 & 0.955 \\
        & LPIPS $\downarrow$ & 0.252 & 0.324 & 0.294 & 0.266 & 0.209 & 0.196 & 0.258 & 0.229 & 0.236 \\
        \hline
        \multirow{3}{*}{ESLAM \cite{johari2023eslam}} 
        & PSNR $\uparrow$ & 29.08 & 25.32 & 27.77 & 29.08 & 33.71 & 30.20 & 28.09 & 28.77 & 29.71 \\
        & SSIM $\uparrow$ & 0.929 & 0.875 & 0.902 & 0.932 & 0.960 & 0.923 & 0.943 & 0.948 & 0.945 \\
        & LPIPS $\downarrow$ & 0.239 & 0.313 & 0.298 & 0.248 & 0.184 & 0.228 & 0.241 & 0.196 & 0.204 \\
        \hline
        \multirow{3}{*}{SplaTAM \cite{keetha2023splatam}} 
        & PSNR $\uparrow$ & 34.11 & 32.86 & 33.89 & 35.25 & 38.26 & 39.17 & 31.97 & 29.70 & 31.81 \\
        & SSIM $\uparrow$ & 0.968 & \secondcolor{0.978} & 0.969 & 0.979 & 0.977 & 0.978 & 0.969 & 0.949 & 0.949 \\
        & LPIPS $\downarrow$ & 0.102 & 0.072 & 0.103 & 0.081 & 0.092 & 0.093 & 0.102 & 0.121 & 0.152 \\
        \midrule
        \multicolumn{11}{c}{\textbf{Semantic SLAM}} \\
        \multirow{3}{*}{SNI-SLAM \cite{zhu2023sni}} 
        & PSNR $\uparrow$ & 29.43 & 25.91 & 28.17 & 29.15 & 31.85 & 30.34 & 29.13 & 28.75 & 30.97 \\
        & SSIM $\uparrow$ & 0.921 & 0.884 & 0.900 & 0.921 & 0.935 & 0.925 & 0.930 & 0.932 & 0.936 \\
        & LPIPS $\downarrow$ & 0.237 & 0.307 & 0.292 & 0.265 & 0.185 & 0.211 & 0.230 & 0.209 & 0.198 \\
        \hline
        \multirow{3}{*}{SGS-SLAM \cite{li2024sgs}} 
        & PSNR $\uparrow$ & 34.66 & 32.50 & \secondcolor{34.25} & 35.10 & \secondcolor{38.54} & \secondcolor{39.20} & \secondcolor{32.90} & \secondcolor{32.05} & 32.75 \\
        & SSIM $\uparrow$ & 0.973 & 0.976 & \secondcolor{0.978} & 0.981 & 0.984 & \secondcolor{0.980} & 0.967 & 0.966 & 0.949 \\
        & LPIPS $\downarrow$ & 0.096 & 0.070 & 0.094 & 0.070 & 0.086 & 0.087 & 0.101 & 0.115 & 0.148 \\
        \hline
        \multirow{3}{*}{SemGauss-SLAM \cite{zhu2024semgauss}} 
        & PSNR $\uparrow$ & \secondcolor{35.03} & \secondcolor{32.55} & 33.32 & \secondcolor{35.15} & 38.39 & 39.07 & 32.11 & 31.60 & \secondcolor{35.00} \\
        & SSIM $\uparrow$ & \cellcolor{green!30}0.982 & \cellcolor{green!30}0.979 & 0.970 & \cellcolor{green!30}0.987 & \cellcolor{green!30}0.989 & 0.972 & \cellcolor{green!30}0.978 & \cellcolor{green!30}0.972 & \cellcolor{green!30}0.978 \\
        & LPIPS $\downarrow$ & \cellcolor{green!30}0.062 & \cellcolor{green!30}0.055 & \cellcolor{green!30}0.054 & \cellcolor{green!30}0.045 & \cellcolor{green!30}0.048 & \cellcolor{green!30}0.046 & \cellcolor{green!30}0.069 & \cellcolor{green!30}0.078 & \cellcolor{green!30}0.093 \\
        \hline
        
        \multirow{3}{*} { \textbf{\MethodName{} (Ours)}}
        & PSNR $\uparrow$ & \bestcolor{35.70} & \bestcolor{32.83} & \bestcolor{34.68} & \bestcolor{36.33} & \bestcolor{39.75} & \bestcolor{40.93} & \bestcolor{33.29} & \bestcolor{32.48} & \bestcolor{35.33} \\
        & SSIM $\uparrow$ & \secondcolor{\textbf{0.980}} & 0.976 & \bestcolor{0.979} & \bestcolor{0.987} & \secondcolor{\textbf{0.988}} & \secondcolor{\textbf{0.989}} & \secondcolor{\textbf{0.975}} & \secondcolor{\textbf{0.971}} & \secondcolor{\textbf{0.976}} \\
        & LPIPS $\downarrow$ & \secondcolor{\textbf{0.067}} & \secondcolor{\textbf{0.060}} & \secondcolor{\textbf{0.063}} & \secondcolor{\textbf{0.052}} & \secondcolor{\textbf{0.050}} & \secondcolor{\textbf{0.049}} & \secondcolor{\textbf{0.083}} & \secondcolor{\textbf{0.081}} & \secondcolor{\textbf{0.094}} \\
        \bottomrule
    \end{tabular}
\end{strip}

\section{Detailed Results on Rendering Quality}
\normalfont
\FloatBarrier

The detailed rendering quality results are presented in \Tab{tab:exp_render_replica}, demonstrating that our method outperforms state-of-the-art approaches.

\end{document}